\documentclass[10pt,twocolumn,letterpaper]{article}

\usepackage[final]{cvpr}      
\usepackage{graphicx}
\usepackage{amsmath}
\usepackage{amssymb}
\usepackage{booktabs}
\usepackage{multirow}
\usepackage{arydshln}
\usepackage{amssymb}
\usepackage{gensymb}
\usepackage{textcomp}

\usepackage[dvipsnames]{xcolor}

\definecolor{cvprblue}{rgb}{0.21,0.49,0.74}
\usepackage[pagebackref,breaklinks,colorlinks,citecolor=cvprblue]{hyperref}

\usepackage[capitalize]{cleveref}
\crefname{section}{Sec.}{Secs.}
\Crefname{section}{Section}{Sections}
\Crefname{table}{Table}{Tables}
\crefname{table}{Tab.}{Tabs.}

\newcommand{\eref}[1]{Eq.~(\ref{#1})}
\newcommand{\tabref}[1]{Table~\ref{#1}}
\newcommand{\figureref}[1]{Figure~\ref{#1}}
\newcommand{\tableref}[1]{Table~\ref{#1}}

\def\paperID{2611} 
\def\confName{CVPR}
\def\confYear{2024}

\begin{document}

%
\title{Deterministic Guidance Diffusion Model for Probabilistic Weather Forecasting} 
\author{Donggeun Yoon\thanks{equal contribution}\\
Chungnam National University\\
{\tt\small ehdrms903@gmail.com}
\and
Minseok Seo\footnotemark[1]\\
SI Analytics\\
{\tt\small minseok.seo@si-analytics.ai}
\and
Doyi Kim\\
SI Analytics\\
{\tt\small doyikim@si-analytics.ai}
\and
Yeji Choi\\
SI Analytics\\
{\tt\small yejichoi@si-analytics.ai}
\and
Donghyeon Cho\thanks{corresponding author}\\
Chungnam National University\\
{\tt\small cdh12242@gmail.com}}
\maketitle
%
%
%
%
\begin{abstract}
Weather forecasting requires not only accuracy but also the ability to perform probabilistic prediction.
However, deterministic weather forecasting methods do not support probabilistic predictions, and conversely, probabilistic models tend to be less accurate.
To address these challenges, in this paper, we introduce the \textbf{\textit{D}}eterministic \textbf{\textit{G}}uidance \textbf{\textit{D}}iffusion \textbf{\textit{M}}odel (DGDM) for probabilistic weather forecasting, integrating benefits of both deterministic and probabilistic approaches.
During the forward process, both the deterministic and probabilistic models are trained end-to-end. 
In the reverse process, weather forecasting leverages the predicted result from the deterministic model, using as an intermediate starting point for the probabilistic model.
By fusing deterministic models with probabilistic models in this manner, DGDM is capable of providing accurate forecasts while also offering probabilistic predictions.
To evaluate DGDM, we assess it on the global weather forecasting dataset (WeatherBench) and the common video frame prediction benchmark (Moving MNIST).
We also introduce and evaluate the Pacific Northwest Windstorm (PNW)-Typhoon weather satellite dataset to verify the effectiveness of DGDM in high-resolution regional forecasting.
As a result of our experiments, DGDM achieves state-of-the-art results not only in global forecasting but also in regional forecasting.
The code is available at: \url{https://github.com/DongGeun-Yoon/DGDM}.
\end{abstract}


\begin{figure}[!t]
    \centering
    \includegraphics[width=0.99\linewidth]{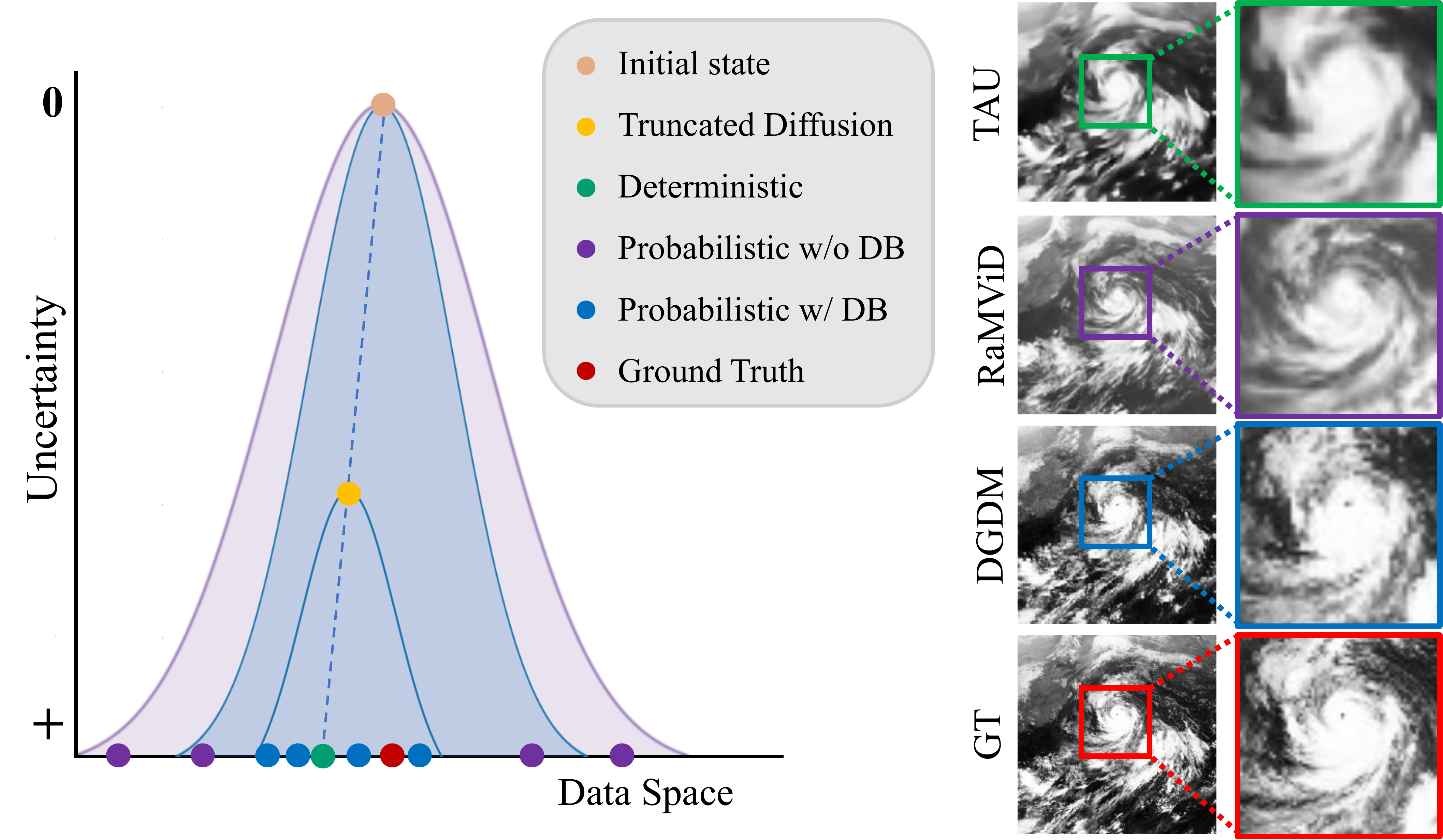}
    \caption{Comparison of uncertainty between probabilistic (RaMViD) and deterministic (TAU) models. The probabilistic model exhibits high diversity, making sample selection near GT challenging. On the other hand, the deterministic model produces a single output, which limits its ability to capture real-world weather variability and contrains its practical application.}
    \label{fig:teaser}
\end{figure}

\section{Introduction}
\label{sec:intro}
Weather is a critical variable that impacts various aspects of daily life, including aviation, logistics, agriculture, and transportation.
To provide accurate weather forecasting, the numerical weather prediction (NWP) method has been predominantly used for most weather forecasting since the 1950s ~\cite{dee2011era, brown2012unified}.
%
NWP employs a simulation-centric framework with vertical and horizontal grids (usually between 10 km and 25 km) that divide the Earth's atmosphere. 
Each grid cell translates the governing atmospheric behavior into partial differential equations (PDEs), which are then solved using numerical integral methods.
Even a single 10-day forecast simulation with the NWP model requires hours of computation on hundreds of supercomputer nodes.
Despite these efforts, phenomena such as turbulent motion and tropical cumulus convection, which occur on a scale of a few kilometers horizontally, cannot be captured by a single deterministic prediction due to being smaller than the grid of NWP models. 
Also, the nonlinearity or randomness of atmospheric phenomena makes it difficult to conduct accurate simulations~\cite{iorio2004effects}.

Therefore, NWP models add small random perturbations to the initial weather conditions and perform multiple simulations to consider the possible scenarios.
%
This ensemble forecasting effectively represents unpredictable events due to nonlinearity or randomness inherent in atmospheric phenomena.
\begin{figure*}[t!]
 \centering
 \includegraphics[width=0.7\linewidth]{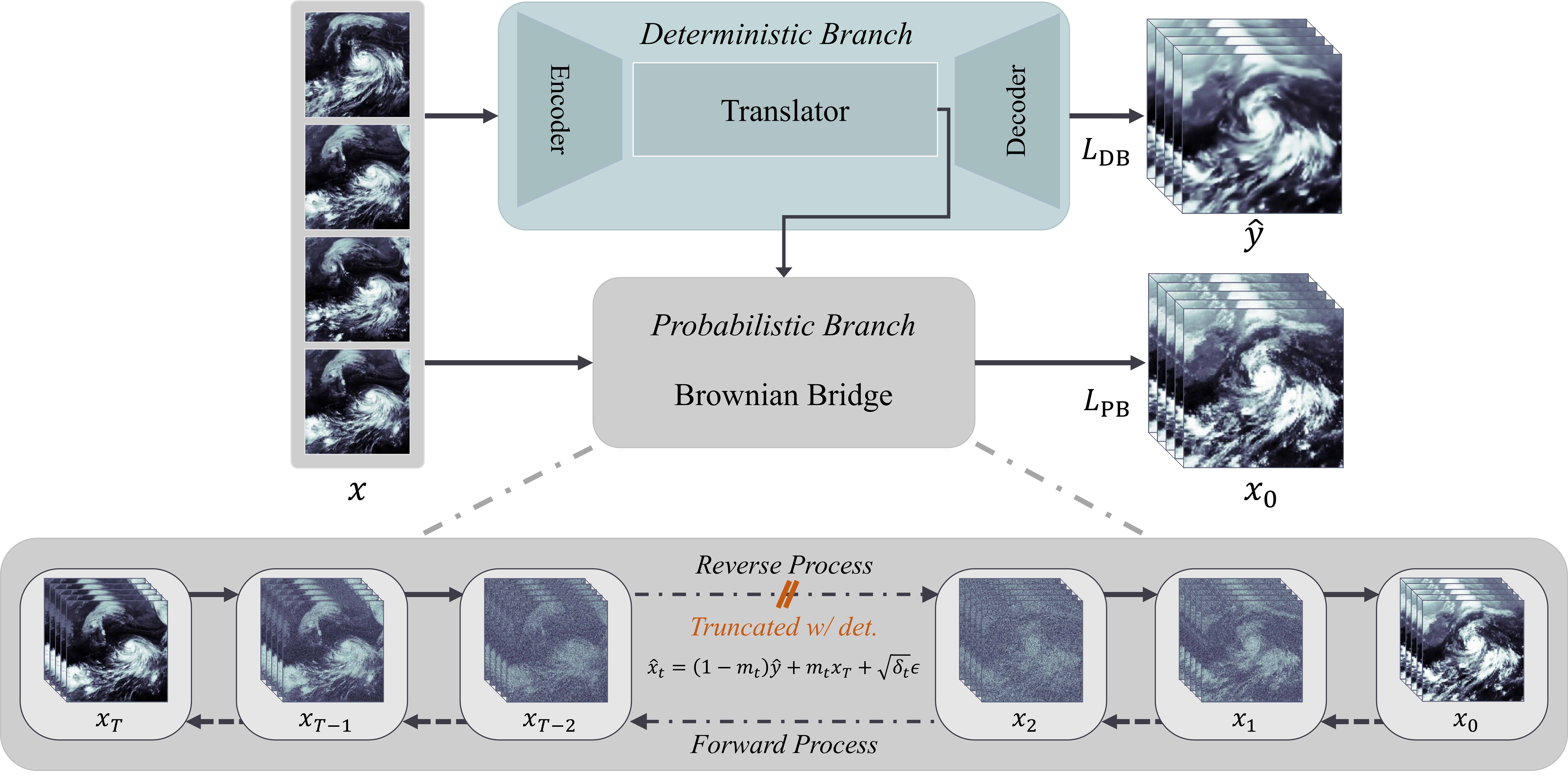}
 \caption{Overview of DGDM: During training, DGDM simultaneously trains both the deterministic and probabilistic models. During inference, the results of the deterministic model are used to truncate the reverse process of the probabilistic model.}
 \label{fig:Overview}
\end{figure*} 
However, it is difficult to configure many ensemble members for the NWP model, which requires a huge amount of computation and time to generate a single prediction.

To alleviate these problems of the NWP model, several data-driven weather forecasting approaches have been proposed~\cite{bi2023accurate, lam2022graphcast}.
While all have surpassed the performance of NWP in 10-day global weather forecasting, they are fundamentally deterministic models and therefore fail to capture the subgrid-scale process\footnote{\textbf{Subgrid-Scale Process}: Atmospheric processes occurring at the smallest scales cannot be represented deterministically within a numerical simulation (\cite{brasseur_jacob_2017}).
}. 
Recently, probabilistic weather forecasting models~\cite{gao2023prediff, leinonen2023latent} using latent diffusion models have been proposed to predict plausible multiple future weather conditions.
They have successfully performed data-driven probabilistic weather forecasting, but due to their diversity, it is unclear which of the various samples should be selected, or ensemble methods may actually degrade performance.
Therefore, data-driven weather forecasting faces a trade-off problem where deterministic models have high performance but cannot perform probabilistic prediction, and probabilistic models generate diverse samples but with lower accuracy.
In this paper, we introduce a \textbf{D}eterministic \textbf{G}uidance \textbf{D}iffusion \textbf{M}odel (DGDM) for probabilistic weather forecasting.
%
As shown in~\Cref{fig:teaser}, weather forecasting using diffusion processes presents challenges, particularly because of the inherent variability in diffusion methods. 
This variability can lead to the selection of samples that deviate significantly from the ground truth. 
On the other hand, deterministic models tend to include all possible futures, resulting in a single and ambiguous outcome.
DGDM addresses the challenges by integrating both deterministic and probabilistic models.
DGDM is structured into two branches: deterministic and probabilistic, as shown in ~\Cref{fig:Overview}.
During the training phase, the deterministic branch takes the initial weather condition, denoted as $ x $, as its input and aims to minimize discrepancies with the future weather condition $ y $.
Simultaneously, the probabilistic branch uses both the initial weather condition $ x = x_{T} $ and the future weather condition $ y = x_{0} $ to train a direct mapping between the domains, using a Brownian bridge stochastic process.
Here, $ T $ represents a diffusion process step.
Subsequently, during the inference phase, the forecast from the deterministic branch serves as an intermediate starting point for the stochastic model, integral to the truncated diffusion process. 
In particular, by adjusting the starting point of the reverse process, DGDM addresses the inherent uncertainty of weather forecasting, allowing it to control the range of possible future weather scenarios.

To verify its efficacy in high-resolution regional weather forecasting, we introduce and evaluate it using the Pacific Northwest Windstorm (PNW)-Typhoon weather satellite dataset.
In addition, We validate DGDM for low-resolution global weather forecasting using the WeatherBench dataset.
Lastly, to determine its effectiveness not only in weather forecasting but also in general video frame prediction, we assess it on the widely used Moving MNIST dataset.
Our evaluation results show that DGDM not only achieves state-of-the-art performance in low-resolution global weather forecasting and high-resolution regional weather forecasting but also demonstrates high performance in the Moving MNIST video frame prediction task.
%

%
%
%
%
%
%
%
%
\section{Related Work}
\label{sec:related}
\subsection{Data-driven Weather Forecasting}
%
%
%
Data-driven weather forecasting is gaining significant attention for providing accuracy comparable to NWP models even without supercomputing resources.
In particular, in many countries where supercomputers cannot be operated, data-driven weather forecasting is a new paradigm that can perform weather forecasting with a single GPU server.

DGMR~\cite{ravuri2021skilful} employed a generative adversarial training approach, utilizing both temporal and spatial discriminators separately.
Shi~\etal~\cite{shi2015convolutional} introduced a ConvLSTM to predict precipitation via autoregressive inference.
Furthermore, Ayzel~\etal~\cite{ayzel2020rainnet} and Trebing~\etal~\cite{trebing2021smaat} presented a data-driven short-term precipitation forecasting leveraging the U-Net architecture~\cite{gruca2022weather4cast,seo2022simple}.
Beyond precipitation, studies such as~\cite{dabrowski2020forecastnet, lam2022graphcast, bi2023accurate} have proposed data-driven models for forecasting global climate variables.
%
The Fourier-based neural network model~\cite{dabrowski2020forecastnet} was designed to generate global data-driven forecasting for atmospheric variables.
Subsequently, models such as GraphCast~\cite{lam2022graphcast} and Pangu-Weather~\cite{bi2023accurate} have outperformed NWP in 10-day forecasts.
While there have been significant advancements, these data-driven forecasting models are deterministic. 
This means that they cannot represent nonlinear and random subgrid-scale phenomena.
\subsection{Video Frame Prediction}
Video frame prediction has various applications such as weather forecasting~\cite{gruca2022weather4cast}, human motion prediction~\cite{ionescu2013human3}, traffic flow prediction~\cite{zhang2017deep} and human robot interaction~\cite{dollar2009pedestrian}.
Video frame prediction is primarily divided into two main categories: autoregressive and non-autoregressive methods.
Traditionally, autoregressive techniques, employing architectures such as ConvLSTM~\cite{shi2015convolutional} and RNN~\cite{wang2017predrnn, wang2018predrnn++, guen2020disentangling}, have been foundational in the field of video frame prediction.
Notably, PhyDNet was introduced in~\cite{guen2020disentangling}, which consists of a two-branch deep architecture designed to disentangle physical dynamics from unknown factors.
%
%
PhyDNet achieved state-of-the-art performance across multiple datasets through a recurrent physical cell (PhyCell) that executes PDE-constrained prediction in a latent space.
%
%
Despite various advancements, autoregressive models inherently suffer from deteriorating performance due to errors accumulation with increasing lead time.

To tackle the problem of error accumulation inherent in autoregressive models, recent research has turned towards non-autoregressive approaches~\cite{gao2022simvp, tan2023temporal, hu2023dynamic, zhong2023mmvp, ning2023mimo, seo2023implicit}.
%
%
For instance, Gao~\etal~\cite{gao2022simvp} designed a non-autoregressive model featuring a multi-in-multi-out structure called SimVP, to prevent error accumulation over specified target lead times.
%
%
%
However, most video frame prediction models were developed based on minimizing the mean squared error (MSE) between predictions and ground truth, which is unsuitable for probabilistic weather forecasting.

\subsection{Video Generation}
%
Video generation technology has made great progress by applying a variety of approaches, including GAN~\cite{luc2020transformation}, VAE~\cite{babaeizadeh2017stochastic, rakhimov2020latent}, and diffusion models~\cite{voleti2022mcvd, hoppe2022diffusion,ho2022video}.
Especially after denoising diffusion probabilistic models (DDPM)~\cite{ho2020denoising}, there has been notable progress in the video generation field.
Ho~\etal~\cite{ho2022video} proposed a video diffusion model (VDM) that generates videos by gradually denoising noisy videos.
VDM is trained to maximize the variational lower bound of the log-likelihood of the data and achieves impressive results on various video-related tasks, such as video generation, editing, and compression.
%
Afterward, RaMViD~\cite{hoppe2022diffusion}, which proposed random-mask video diffusion, and MCVD~\cite{voleti2022mcvd}, which integrated video prediction, generation, and interpolation into one framework, were proposed.
While these probabilistic models can generate plausible future frames, they are not optimized for accurate predictions.
Therefore, simply applying video generation models to weather forecasting is unsuitable for expressing uncertainty because they cannot control their diversity.
The proposed DGDM addresses this issue by effectively merging probabilistic and deterministic models, offering a probabilistic model that can still control the range of possible future weather scenarios.
%
%
%
%
\section{Method}
\label{sec:method}
The objective of DGDM is to forecast future weather condition $y\in \mathbb R^{C \times \hat{L}\times H\times W}$ given the initial weather condition $x\in \mathbb R^{C \times L\times H\times W}$.
Where $ H $, $ W $, and $ C $ represent the height, width, and channels, respectively.
$ L $ and $ \hat{L} $ denote the lengths of the observed and forecasted frames.
As shown in~\Cref{fig:Overview}, DGDM consists of two main components: the deterministic branch and the probabilistic branch.
The probabilistic branch accounts for the inherent uncertainties in future forecasting, ensuring a wide range of possible futures.
On the other hand, the deterministic branch improves the accuracy of forecasts.
Furthermore, the results from the deterministic branch are used to modulate uncertainties, which in turn helps to control the diversity of potential future weather scenarios in the probabilistic branch.
\subsection{Deterministic Branch}
The deterministic branch (DB) of the DGDM adopts a non-autoregressive structure, which is beneficial for achieving high performance as it prevents error accumulation at fixed lead times.
The model architecture consists of an encoder $ e(\cdot) $, a translator $ st(\cdot) $, and a decoder $ d(\cdot) $, similar to the structure utilized in TAU~\cite{tan2023temporal}.
Given an input $ x $, the loss function of the deterministic branch is formulated as:
\begin{equation}
    L_{DB} = \lVert y - d(st(e(x))) \rVert^{2}.
\end{equation}

\subsection{Probabilistic Branch}
The probabilistic forecasting method in NWP is to apply different small random perturbations to the initial conditions each time the model is run, thereby generating diverse outcomes.
Inspired by the NWP process, which produces probabilistic predictions through deterministic start and end points coupled with perturbations to the initial conditions, the probabilistic branch of the DGDM has been designed.
\paragraph{Brownian Bridge Diffusion Process} When we consider the problem as one of predicting the stochastic trajectories between an initial weather condition $ x $ and a future weather condition $ y $, a Brownian bridge can be applied as a continuous-time stochastic model where the probability distribution during the diffusion process is conditional on the starting and ending states.
Specifically, the state distribution at each time step of a Brownian bridge process starting from point $ y = x_0 $ with $ x_0 \sim q_{\text{data}}(x_0) $ at $ t = 0 $ and ending at point $ x = x_T $ at $ t = T $ can be formulated as:
\begin{equation}
p(x_t | x_0, x_T) = \mathcal{N}\left((1 - \frac{t}{T})x_0 + \frac{t}{T}x_T, \frac{t(T - t)}{T}\mathbf{I}\right),
\end{equation}
where $ \mathcal{N} $ denotes the normal distribution and $ \mathbf{I} $ is the identity matrix that scales the variance of the distribution.

We can define the Brownian bridge diffusion process in the simplified notation of DDPM as follows:
\begin{equation}
    \begin{array}{c}
        q(x_{t}|x_{0},x_T)=N(x_{t};(1-m_{t})x_{0}+m_{t}x_T,\delta_{t} I), \\
        x_{0}=y, m_{t}=\cfrac{t}{T}.
    \end{array}
    \label{eq:bbdm_forward}
\end{equation}
\begin{figure}[t!]
 \centering
 \includegraphics[width=1.0\linewidth]{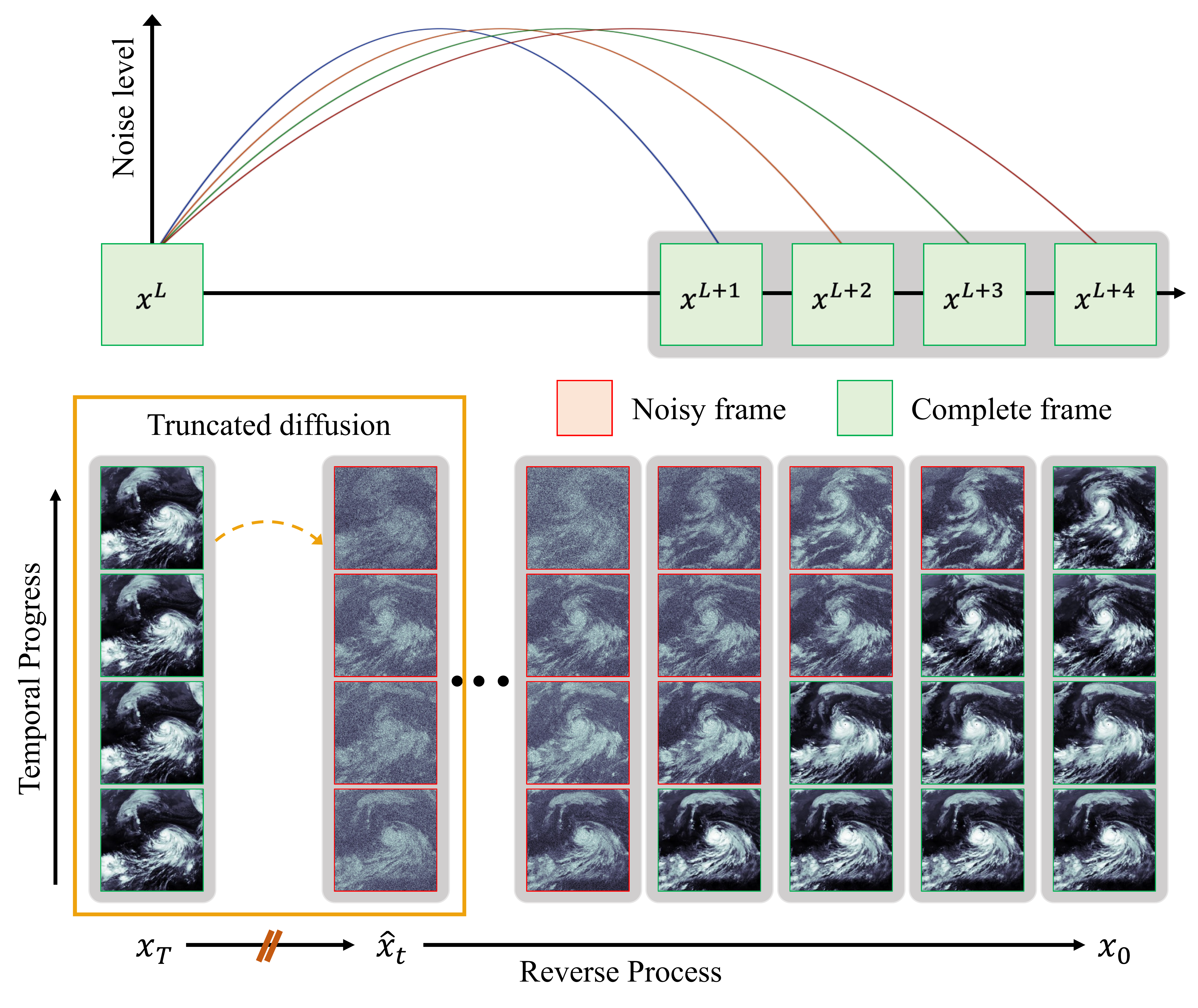}
 \caption{
The reverse process of DGDM employs the sequential variance schedule (SVS) and truncated diffusion. SVS allocates varying diffusion steps according to lead time; shorter lead times are assigned fewer diffusion steps, while longer lead times linearly receive more steps. Truncated diffusion starts the reverse process at the intermediate state by utilizing deterministic result.
}
 \label{fig:SVS}
\end{figure}
Note that, as in the case of image to image translation tasks where the Brownian bridge diffusion process has been successfully applied~\cite{li2023bbdm}, in DGDM, $ \delta_{t} $ is defined as $ 2(m_{t} - m_{t}^{2}) $.
\paragraph{Deterministic Guidance Diffusion Model for Probabilistic Weather Forecasting}
The structure of the probabilistic branch (PB) of DGDM is modeled after VDM~\cite{ho2022video}.
Unlike VDMs that generate video clips from noise, PB applies ~\eref{eq:bbdm_forward} when given input $ x $ and output $ y $.
However, using input $ x $ directly leads to two main issues.
First, the length of input $ L $ must match the target length $ \hat{L} $ for the diffusion model to function properly, which complicates the control of output length $ \hat{L} $.
%
Second, $ x^{0} $, the first frame of the input, has the greatest lead time distance from the first frame of the target $ y^{0} $.
In contrast, $ x^{L} $ is closer to $ y^{0} $, where the superscript number denotes the specific frame.
This makes the direct use of $ x $ to perform ~\eref{eq:bbdm_forward} inefficient.
To address these issues, we replicate $ x^{L} $ $ \hat{L} $ times to serve as the input for PB, which we denote as $ x^{\hat{L}} $.
Consequently, the forward process of PB, incorporating ~\eref{eq:bbdm_forward}, is defined as follows. %
\begin{equation}
    \def\arraystretch{1.4}
    \begin{array}{c}
        x_{t}=(1-m_{t})x_{0} + m_{t}x^{\hat{L}} + \sqrt{\delta_{t}}\epsilon, \\
        x_{t-1}=(1-m_{t-1})x_{0} + m_{t-1}x^{\hat{L}} + \sqrt{\delta_{t-1}}\epsilon.
    \end{array}
    \label{eq:bbdm_inter}
\end{equation}
Here, $m_{t}$ stands for a time weight, linearly increasing from 0 to 1 with respect to $t$. 
Concurrently, $\delta_{t}$ denotes the variance defined as $2(m_{t}-m_{t}^{2})$, and $\epsilon$ corresponds to Gaussian noise, specifically $N\sim(0,1)$.
However, as $ x^{\hat{L}} $ consists of identical frames, spatial-temporal modeling is not feasible.
To resolve this issue, we employ the feature $ z $ extracted from $ st(e(x)) $ in the DB to perform cross-attention~\cite{rombach2022high} as a condition in PB.
Therefore, the object function of PB is as follows.

\begin{equation}
    L_{\text{PB}} = \mathbb{E} \left| \left| m_t(x^{\hat{L}} - x_{0}) + \sqrt{\delta_t} \epsilon - \epsilon_{\theta}(x_t, t, z) \right| \right|^2.
\label{eq:train_normal}
\end{equation}

The DB and PB are jointly trained in an end-to-end manner, and the total objective function for DGDM is defined as follows:
\begin{equation}
    L_{\text{total}} =  L_{\text{PB}} + L_{\text{DB}}.
\label{eq:train_normal}
\end{equation}
\begin{table*}[t!]
\center
\resizebox{1.5\columnwidth}{!}{%
\begin{tabular}{c|c|c|c|ccccc}
\hline \hline
\multirow{2}{*}{\textbf{Model}} & \multirow{2}{*}{\textbf{Diversity}} & \multirow{2}{*}{\textbf{\#Param. (M)}} & \multirow{2}{*}{\textbf{\#Flops (G)}} & \multicolumn{5}{c}{\textbf{Evaluate metric}} \\ \cline{5-9} 
                       &                            &                               &                              & MAE$\downarrow$ & MSE $\downarrow$ & PSNR $\uparrow$ & SSIM $\uparrow$ & FVD $\downarrow$  \\ \hline
ConvLSTM~\cite{shi2015convolutional} & $\times$        &  15.0M              & 56.79G                      &  90.63         & 29.79           & 22.14          & 0.928          & 79.193    \\
PredRNN~\cite{wang2017predrnn}       & $\times$        &  23.8M              & 0.116T                      &  72.82         & 23.96           & 23.28          & 0.946          & 50.407     \\
PredRNN++~\cite{wang2018predrnn++}   & $\times$        &  38.6M              & 0.172T                      &  69.58         & 22.05           & 23.65          & 0.950          & 45.731    \\
MIM~\cite{wang2019memory}            & $\times$        &  38.0M              & 179.2G                      &  70.67         & 22.92           & 23.53          & 0.948          & 47.530    \\
PhyDNet~\cite{guen2020disentangling} & $\times$        &   3.1M              & 15.3G	                   &  61.47         & 20.35           & 24.21          & 0.955          & 38.752     \\
SimVP~\cite{gao2022simvp}            & $\times$        &  58.0M              & 9.431G                      &  89.04         & 32.14           & 21.83          & 0.926          & 72.969     \\
TAU~\cite{tan2023temporal}           & $\times$        &  46.8M              & 15.953G                     &  51.46         & \textbf{15.68}  & \textbf{25.71} & \textbf{0.966} & 28.169      \\ \hline
VDM~\cite{ho2022video}               & \checkmark      &  35.71M             & 77.446G                     & 123.12         & 86.33           & 18.71          & 0.879          &  8.800      \\
MCVD~\cite{voleti2022mcvd}           & \checkmark      &   28.0M             & 9.915G $\times$ 100         & 172.47         & 64.68           & 19.23          & 0.565          &  8.161   \\ 
RaMViD~\cite{hoppe2022diffusion}     & \checkmark      & 235.058M            & 1.052T $\times$ 1000        & 123.76         & 81.26           & 18.87          & 0.878          & 12.059 \\ \hline
DGDM-DB                              & $\times$        &   27.0M             & 11.456G                     & 56.45          & \underline{17.88} & 24.94          & 0.961          & 19.216 \\
DGDM-SB                              & \checkmark      &   63.3M             & 77.288G $\times$ 100        & 50.21          & 20.96           & 25.08          & \underline{0.962} & \underline{7.461} \\
DGDM-Best                            & $\times$        &   63.3M             & 77.288G $\times$ 100        & \textbf{47.31} & 19.14           & \underline{25.59} & \textbf{0.966} & \textbf{7.427} \\
DGDM-Average                         & \checkmark      &   63.3M             & \textbf{(}77.288G $\times$ 100\textbf{)} $\times$ 10 & \underline{48.54}        & 20.52           & 25.22          & \textbf{0.966} &  9.617 \\ \hline \hline
\end{tabular}%
}
\caption{Performance comparison results from the Moving MNIST dataset. For diffusion models, the amount of computation is indicated by the number of steps required in the reverse process. \textbf{Bold} indicates the best performance, and \underline{underline} indicates the second-best performance.}
\label{tab:MMNIST-results}
\end{table*}
\paragraph{Sequential Variance Schedule}
In weather forecasting, the uncertainty increases with the length of the lead time.
To capture this characteristic, we propose the sequential variance schedule (SVS).
SVS is utilized during both the training and inference phases. Given the total number of diffusion steps $ T $ and the output length $ \hat{L} $, the SVS for each lead time is defined by the equation:
\begin{equation}
     \textrm{SVS} = \{ T - (\hat{L} - i) \cdot S : i = 1, …, \hat{L}  \}
\end{equation}
Here, $S$ denotes the step size of the diffusion steps.
\subsection{Inference}
To forecast the future weather condition $y$, the reverse process of the Brownian bridge starts at $x_{T} = x^{\hat{L}}$.
Using the trained model, we then proceed through the reverse process that incrementally moves from $x_t$ to $x_{t-1}$.
\begin{equation}
    p_{\theta}(x_{t-1}|x_{t}) = N(x_{t-1};\mu_{\theta}(x_{t},t, z), \tilde{\delta}_{t} I),
\end{equation}
where $\mu_{\theta}(x_{t},t,z)$ denotes the estimated mean value of the noise, and $\tilde{\delta}_{t}$ is the variance of noise.
We reach the ending point of the diffusion process, where $x_0=y$.
For acceleration, the inference process integrates a non-Markovian chain, as detailed in BBDM~\cite{li2023bbdm}.
\begin{table*}[t!]
\centering
\resizebox{2\columnwidth}{!}{%
\begin{tabular}{c|ccc|ccc|ccc|ccc|ccc}
\hline \hline
\multirow{3}{*}{\textbf{Model}}  & \multicolumn{15}{c}{\textbf{PNW-Typhoon}}      \\ \cline{2-16} 
  & \multicolumn{3}{c|}{MSE $\downarrow$}          & \multicolumn{3}{c|}{MAE $\downarrow$}          & \multicolumn{3}{c|}{PSNR $\uparrow$}         & \multicolumn{3}{c|}{SSIM $\uparrow$}         & \multicolumn{3}{c}{FVD $\downarrow$} \\ \cline{2-16} 
                & IR & SW & \multicolumn{1}{c|}{WV} & IR & SW & \multicolumn{1}{c|}{WV} & IR & SW & \multicolumn{1}{c|}{WV} & IR & SW & \multicolumn{1}{c|}{WV} & IR     & SW     & WV    \\ \hline
ConvLSTM~\cite{shi2015convolutional} & 937.80 & 1075.43 & 409.65   & 2973.50   & 3221.82   & 1952.39   & 13.06   & 14.31   & 17.94  & 0.399   & 0.400   & 0.642  & 1689.39 & 1321.36 & 993.59   \\
PredRNN~\cite{wang2017predrnn}       & 774.73 & 804.28  & 253.28   & 2342.76   & 2933.85   & 1756.32   & 14.23   & 14.57   & 18.34  & 0.401   & 0.403   & 0.629  & 1640.11 & 1293.09 & 1033.27  \\
PredRNN++~\cite{wang2018predrnn++}   & 667.18 & 739.01  & 213.51   & 2372.64   & 3088.18   & 1552.26   & 13.51   & 14.55   & 20.29  & 0.414   & 0.410   & 0.630  & 1340.61 & 1040.10 & 926.51   \\
MIM~\cite{wang2019memory}            & 625.99 & 683.51  & 195.14   & 2398.68   & 3127.97   & 1567.28   & 13.85   & 13.95   & 17.55  & 0.392   & 0.392   & 0.613  & 1262.59 & 969.19  & 954.53   \\
PhyDNet~\cite{guen2020disentangling} & 655.14 & 728.93  & 189.02   & 2460.73   & 3144.94   & 1675.00   & 14.42   & 14.10   & 18.96  & 0.408   & 0.398   & 0.576  & 1305.54 & 1007.79 & 944.65   \\
SimVP~\cite{gao2022simvp}            & 643.06 & 706.72  & 204.33   & 2452.73   & 3079.17   & 1609.57   & 14.01   & 14.21   & 19.17  & 0.401   & 0.406   & 0.598  & 1283.45 & 991.53  & 940.42   \\
TAU~\cite{tan2023temporal}           & 565.47 & 664.50  & 166.11   & 2117.84   & 2842.39   & 1493.33   & 15.12   & 14.82   & 19.49  & 0.404   & 0.418   & 0.603  & 997.47  & 880.41  & 891.90   \\ \hline
VDM\cite{ho2022video}                & 881.08 & 1128.37 & 401.52   & 2794.82   & 3355.41   & 2063.66   & 13.43   & 12.31   & 17.07  & 0.371   & 0.383   & 0.607  & 830.29  & 727.09  & 573.47   \\
MCVD\cite{voleti2022mcvd}            & 605.51 & 904.93  & 472.19   & 2166.98   & 2799.92   & 2300.42   & 14.89   & 13.57   & 16.74  & 0.430   & 0.433   & 0.647  & 737.99  & 439.97  & 481.70   \\ 
RaMViD\cite{hoppe2022diffusion}      & 770.72 & 1152.08 & 341.68   & 2781.69   & 3355.09   & 1803.71   & 13.96   & 12.55   & 18.39  & 0.392   & 0.392   & 0.624  & 1091.97 & 935.57  & \textbf{395.73}   \\ \hline    
DGDM-DB                              & 495.52 & 576.78  & 133.16   & 1949.46   & 2162.01   & 1031.67   & 15.89   & 15.43   & 21.785 & 0.442   & 0.436   & 0.653  & 819.72  & 710.32  & 799.30   \\
DGDM-SB                              & 461.90 & 540.79  & 121.27   & 1875.08   & 2065.37   & 1003.18   & 16.08   & 15.57   & 21.910 & 0.478   & 0.457   & 0.678  & \underline{705.56}  & \underline{422.03}  & 508.36   \\
DGDM-Best                            & \underline{459.85} & \underline{533.35}  & \underline{120.75}   & \underline{1871.08}   & \underline{2047.80}   & \underline{1000.82}   & \underline{16.10}   & \underline{15.63}   & \underline{21.932} & \underline{0.480}   & \underline{0.460}   & \underline{0.679}  & \textbf{698.93}  & \textbf{414.07}  & \underline{502.26}   \\
DGDM-Average                         & \textbf{456.11} & \textbf{516.66}  & \textbf{120.34}   & \textbf{1863.26}   & \textbf{2004.7}1   & \textbf{998.31}    & \textbf{16.15}   & \textbf{15.83}   & \textbf{21.955} & \textbf{0.482}   & \textbf{0.483}   & \textbf{0.682}  & 733.03  & 482.80  & 531.14   \\ \hline \hline
\end{tabular}%
}
\caption{Performance comparison results from the PNW-Typhoon dataset. IR stands for infrared red, WV stands for water vapor, and SW stands for short wave, representing each channel of the weather observation satellite.}
\label{tab:PNW-Typhoon_main}
\end{table*}
\paragraph{Truncated Diffusion with Deterministic}
By incorporating the results of DB $ \hat{y} $ into the reverse process, we can truncate the diffusion process, as illustrated in~\Cref{fig:SVS}.
Specifically, substituting $ \hat{y} $ for $ x_0 $ in~\eref{eq:bbdm_inter} allows us to begin from an intermediate state, $ \hat{x_t} $, rather than starting directly from the initial condition.
The truncated reverse diffusion process of DGDM follows the equation:
\begin{equation}    
    \hat{x}_{t} = (1 - m_t)\hat{y} + m_t x_T + \sqrt{\delta_t} \epsilon.
    \label{eq:truncated}
\end{equation}
Truncated diffusion not only modulates the diversity of results but also accelerates the speed of inference.

\section{Experiments}
\label{sec:exp}
In this section, we provide a comparative analysis using synthetic dataset and weather forecasting benchmarks to demonstrate the effectiveness of DGDM.
Furthermore, we conduct comprehensive ablation studies to demonstrate the contributions of the components within our approach.
\subsection{Experimental Setting}
\paragraph{Dataset} We use the Moving MNIST dataset, which is frequently used in future frame prediction tasks.
Additionally, experiments are performed on two weather datasets: PNW-Typhoon and Weatherbench dataset.
In our experiments, we train on the training set of each dataset and evaluate it on the test set.
%
\begin{itemize}
    \item \textbf{Moving MNIST~\cite{srivastava2015unsupervised}} The Moving MNIST dataset contains 10,000 video sequences, each consisting of 20 frames. In each video sequence, two digits move independently around the frame, which has a spatial resolution of $64 \times 64$ pixels. The digits frequently intersect with each other and bounce off the edges of the frame.  Consequently, the values of $C,L,\hat{L},H,W$ are 1, 10, 10, 64, and 64 respectively.
    \item \textbf{PNW-Typhoon} The Pacific Northwest Windstorm (PNW)-Typhoon weather satellite dataset is a collection that records typhoons observed from January 2019 to September 2023. The PNW-Typhoon dataset is sourced from the GK2A (GEO-KOMPSAT-2A) satellite. PNW-Typhoon dataset used observations at 1-hour intervals over the East Asia regions with a 2 km spatial resolution.
    All data preprocessing strictly adhered to the official GK2A user manual\footnote{\url{https://nmsc.kma.go.kr/enhome/html/base/cmm/selectPage.do?page=satellite.gk2a.intro}} to maintain the integrity of the GK2A physical value data. The image dimensions in the GK2A dataset are $128 \times 128$. In our experiments, we exclusively use channels of the infrared ray (IR) at 10.5 $\mu$m, short wave infrared ray (SW) at 0.38 $\mu$m, and water vapor (WV) at 0.69 $\mu$m. Typhoons from 2019 to 2022 are used for training, while the ones from 2023 are used for testing. Consequently, the values for $C, L, \hat{L}, H, W$ are 3, 10, 10, 128, and 128, respectively.
    \item \textbf{Weatherbench~\cite{rasp2020weatherbench}} The WeatherBench dataset is a comprehensive weather forecasting collection that encompasses various climatic factors.
    The raw data is re-gridded to a $5.625^\circ$ resolution, equivalent to a $32 \times 64$ grid. 
    We employ the WeatherBench-S framework from~\cite{tan2023openstl}, where each climatic factor is trained individually.
    The model is trained using data spanning 2010 to 2015, validated using data from 2016, and tested with 2017-2018 data, maintaining a temporal granularity of one hour.
    Consequently, the values for $C, L, \hat{L}, H, W$ are 1, 12, 12, 32, and 64, respectively.
\end{itemize}

\paragraph{Evaluation Metric}
For evaluating DGDM, we use mean squared error (MSE) to heavily penalize larger errors and mean absolute error (MAE) for a linear error penalty. 
We utilize peak signal-to-noise ratio (PSNR) to evaluate the quality of signal representation against corrupting noise, and structural similarity index measure (SSIM)~\cite{wang2004image} to assess perceptual results. 
Additionally, we employ Frechet video distance (FVD) to measure the similarity between the generated videos and the ground truth videos in feature space. 

\paragraph{Implementation Details}
For our implementation, the Adam optimizer is employed with a learning rate of $1e^{-4}$ for the diffusion model and $3e^{-4}$ for the deterministic model.
We train with a mini-batch size of 8.
The learning rate is adaptively adjusted using the ReduceLROnPlateau scheduler.
We adopt a forwarding process of 1,000 steps and a reverse process of 200 steps. 
However, the reverse process is truncated after 100 steps.
To enhance model stability, we implement exponential moving average (EMA) with a decay factor of 0.995.
The number of training epochs varies by dataset: 2,000 for the Moving MNIST, 200 for the PNW-typhoon dataset, and 50 for the WeatherBench dataset.
All experiments are conducted using the PyTorch framework on a single A100 GPU.
Detailed network architectures are provided in the supplementary material.
\begin{figure}[t]
\renewcommand{\arraystretch}{0.4}
    \centering
    \begin{tabular}{c}
    \includegraphics[width=0.8\linewidth]{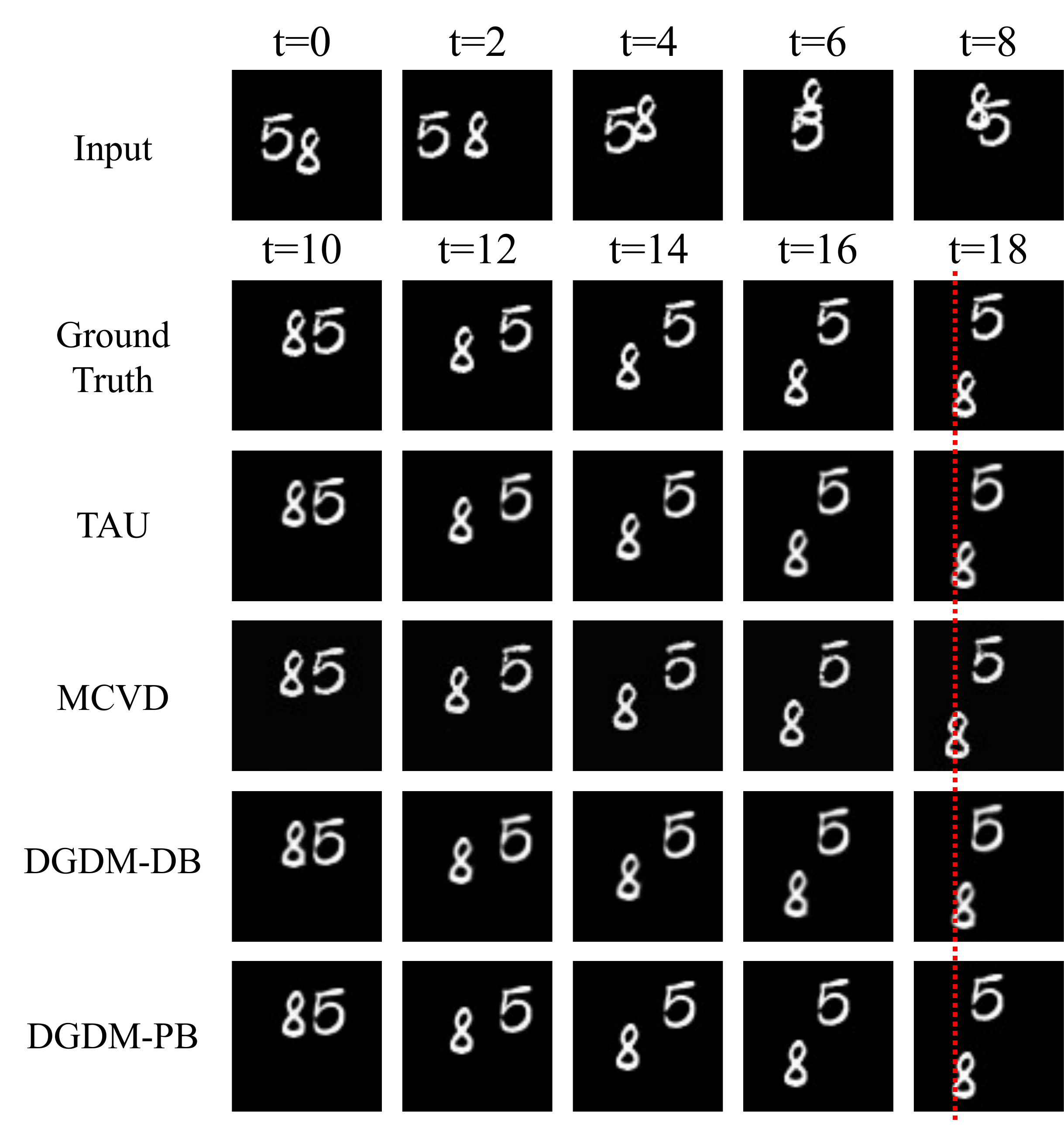}
    \end{tabular}
\caption{Visualization of Moving MNIST with time stride of 2.}
\label{fig:MMNIST}
\end{figure}
\begin{figure*}[t!]
 \centering
 \includegraphics[width=0.8\linewidth]{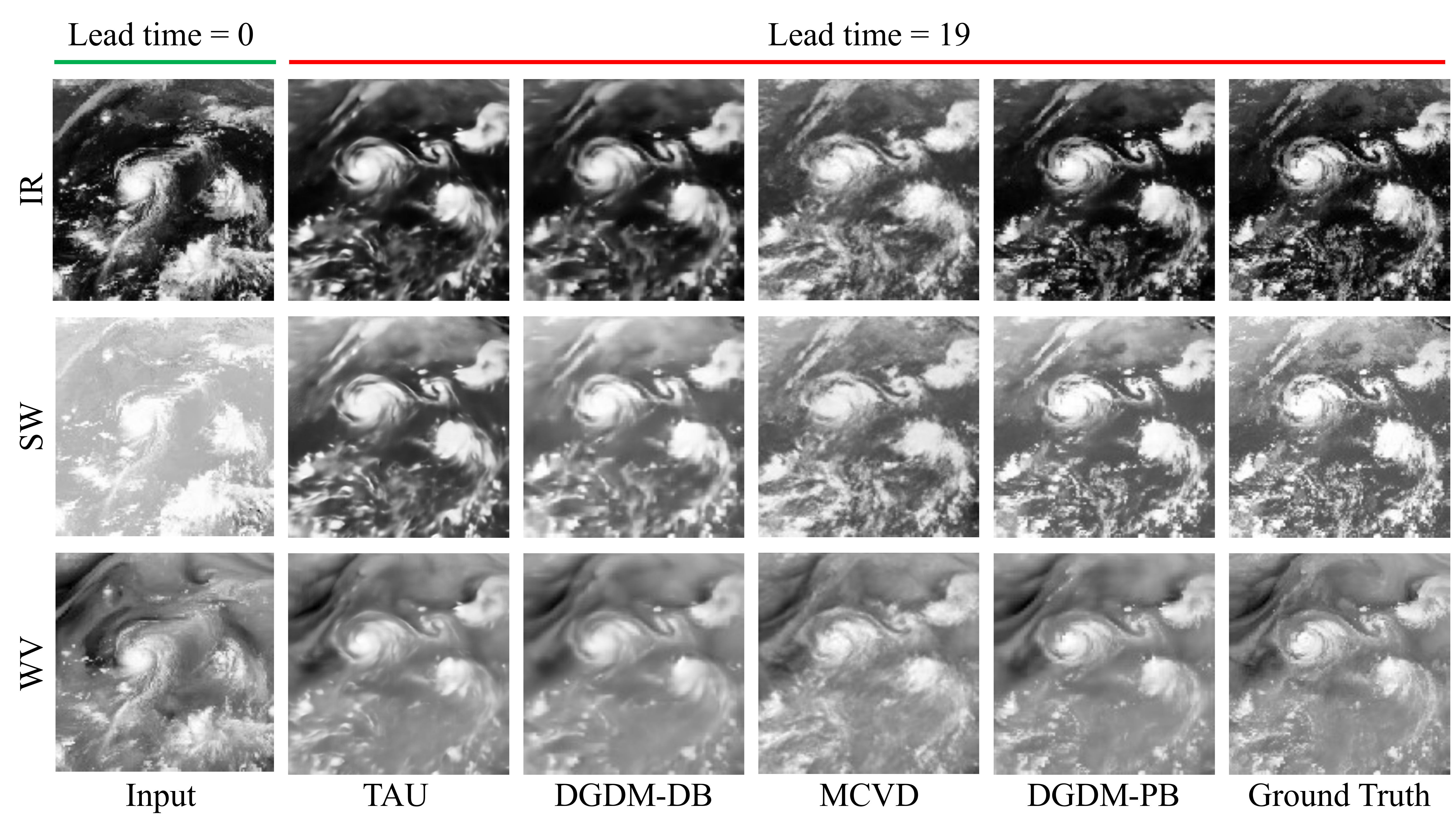}
 \caption{Qualitative comparison results on the PWN-Typhoon dataset. TAU~\cite{tan2023temporal} and MCVD~\cite{voleti2022mcvd} are the best-performing approaches in deterministic and probabilistic models, respectively.}
 \label{fig:pwn_typhoon}
\end{figure*}
\subsection{Experimental Results}
\paragraph{Moving MNIST} We evaluate DGDM on the Moving MNIST dataset to verify its effectiveness in the video frame prediction task.
We compare the performance with existing data-driven video frame prediction models~\cite{shi2015convolutional, wang2017predrnn,wang2018predrnn++, wang2019memory, guen2020disentangling, gao2022simvp, tan2023temporal} on the Moving MNIST dataset.
As shown in~\Cref{tab:MMNIST-results}, models without diversity all achieve better scores in MAE, MSE, PSNR, and SSIM, but exhibit low FVD.
This suggests that while deterministic models are adept at predicting future movements in the Moving MNIST dataset, they struggle to replicate finer details, such as texture.
In contrast, video generation models with diversity score competitively on FVD yet have significantly worse MAE, MSE, PSNR, and SSIM compared to deterministic models.
As illustrated in ~\figureref{fig:MMNIST}, these experimental analyses offer insights that corroborate our observations: deterministic models precisely predict locations but lack the clarity of the GT, whereas probabilistic models, despite their clarity, do not predict locations as accurately.
However, the DGDM demonstrates a close resemblance to the GT in both location accuracy and clarity.

Despite the existence of diversity, DGDM is able to obtain state-of-the-art results for all MAE, SSIM, and FVD.
These experimental results show that DGDM is a model that is as accurate as we intended but produces diverse results.
Furthermore, our model possesses diversity, making ensemble methods possible. 
When ensemble techniques are employed, DGDM achieves better scores in MAE, MSE, PSNR, and SSIM.
However, despite these achievements, performance with respect to FVD declined in random selection. 
The same averaging might result in outputs that, while closer to the GT in a pixel-wise sense, might lack some of the higher-level semantic nuances or temporal coherence.
\paragraph{PNW-Typhoon}
\tableref{tab:PNW-Typhoon_main} shows the results of verifying video frame prediction methods and video generation methods on the PNW-Typhoon dataset.
The experimental results in ~\tableref{tab:PNW-Typhoon_main} show that, unlike the experimental results in ~\tableref{tab:MMNIST-results}, most of the deterministic models perform poorly in MSE and MAE compared to probabilistic models.
These experimental results show that probabilistic models have potential in the field of high-resolution regional weather forecasting.
\begin{table}[t!]
\resizebox{\columnwidth}{!}{%
\begin{tabular}{c|cccccc}
\hline \hline
\multirow{3}{*}{\textbf{Model}} & \multicolumn{6}{c}{\textbf{WeatherBench}}                                                                          \\ \cline{2-7} 
                       & \multicolumn{2}{c|}{Temperature (t2m)} & \multicolumn{2}{c|}{Humidity (r)} & \multicolumn{2}{c}{Wind (uv10)} \\ 
                       &     MSE    & \multicolumn{1}{c|}{RMSE}        &  MSE     & \multicolumn{1}{c|}{RMSE}     &      MSE          &    RMSE            \\ \hline
                       ConvLSTM~\cite{shi2015convolutional} &     1.521    & \multicolumn{1}{c|}{1.233}        &  35.146     & \multicolumn{1}{c|}{5.928}     &    1.8976            &     1.3775           \\
                   PredRNN~\cite{wang2017predrnn}    &  1.331       & \multicolumn{1}{c|}{1.154}        &   37.611    & \multicolumn{1}{c|}{6.133}     &  1.8810              &   1.3715             \\
                   PredRNN++~\cite{wang2018predrnn++}    &   1.634      & \multicolumn{1}{c|}{1.278}        &  35.146	     & \multicolumn{1}{c|}{5.928}     &  1.8727              &   1.3685             \\                   
                   MIM~\cite{wang2019memory}    &   1.784      & \multicolumn{1}{c|}{1.336}        &   36.534    & \multicolumn{1}{c|}{6.044}     &    3.1399            &   1.7720             \\                   
                   SimVP~\cite{gao2022simvp}    &     1.238    & \multicolumn{1}{c|}{1.113}        &   34.355    & \multicolumn{1}{c|}{5.861}     &   1.9993             &   1.4140             \\
                   TAU~\cite{tan2023temporal}    &   \underline{1.162}      & \multicolumn{1}{c|}{\underline{1.078}}        &   31.831    & \multicolumn{1}{c|}{5.642}     &   \underline{1.5925}             &    \underline{1.2619}            \\   \hline   
                   VDM~\cite{ho2022video}    &  2.343       & \multicolumn{1}{c|}{1.530}        &  43.293     & \multicolumn{1}{c|}{6.579}     &     2.2345           &      1.4948          \\
                   MCVD~\cite{voleti2022mcvd}    &  2.512       & \multicolumn{1}{c|}{1.584}        &  45.691     & \multicolumn{1}{c|}{6.759}     &      2.2213          &   1.4904             \\
                   RaMViD~\cite{hoppe2022diffusion}    & 1.908        & \multicolumn{1}{c|}{1.381}        &    39.028   & \multicolumn{1}{c|}{6.247}     &       2.7639         &   1.6624             \\ \hline
                  DGDM-Best  &    \textbf{1.025}     & \multicolumn{1}{c|}{\textbf{1.012}}        &   \textbf{28.572}    & \multicolumn{1}{c|}{\textbf{5.345}}     &   \textbf{1.5914}             &    \textbf{1.2615}            \\ 
                  DGDM-Average    & 1.183        & \multicolumn{1}{c|}{1.087}        &  \underline{30.326}     & \multicolumn{1}{c|}{\underline{5.506}}     &       1.6439         &    1.2821            \\ 
                   \hline \hline
\end{tabular}%
}
\caption{Quantitative comparison results between DGDM and other baselines on the WeatherBench dataset.}
\label{tab:weatherbench}
\end{table}

\figureref{fig:pwn_typhoon} presents a qualitative analysis of deterministic and probabilistic models, specifically, TAU and MCVD.
In typhoon analysis, aspects like the eye of the typhoon, its size, and the way clouds disperse and coalesce are critical.
As depicted in~\figureref{fig:pwn_typhoon}, while the deterministic models simulate the size of the typhoon well, the results are too blurry to definitively assess the eye of the typhoon or the cloud patterns.
On the other hand, MCVD provides clearer images but falls short in predicting the size of the typhoon and cloud formation.
DGDM, a probabilistic model, most accurately simulates the size of a typhoon and is also the most precise in predicting the formation and dissipation of clouds.
These experimental results suggest that DGDM is highly effective for high-resolution regional weather forecasting.

\paragraph{WeatherBench}
\tabref{tab:weatherbench} shows the results of quantitative experiments in WeatherBench, a global weather forecasting dataset.
DGDM-Best, a method of selecting the sample with the highest performance among all 20 samples, achieves the best performance in the WeatherBench dataset.
In addition, DGDM-Average, which ensembles all 20 models, shows the second-best performance in the Humidity modality.
%
%
These experimental results indicate that DGDM has high usability even if a single output is not selected and simply ensembled with average.

\subsection{Ablation Study}
We perform ablation studies on the Moving MNIST dataset to evaluate the effects of the components and the initial point.
\vspace{-4mm}
\begin{table}[t!]
    \centering
    \resizebox{\columnwidth}{!}{%
    \begin{tabular}{ccccc|cc|cc} \hline \hline
    \multicolumn{5}{c}{\textbf{Components}}                            & \multicolumn{2}{|c|}{\textbf{Deterministic}} & \multicolumn{2}{c}{\textbf{Probabilsitic}}  \\ \hline
    DB         & PB          & BB         & LF          & SVS          & MAE           & FVD    & MAE        & FVD      \\ \hline
    \checkmark &             &            &             &              & 58.13         & 18.50  & -          & -        \\ 
               & \checkmark  &            &             &              & -             & -      & 123.20     & 8.80     \\ \hline
    \checkmark & \checkmark  &            &             &              & 95.09         & 58.83  & 110.07     & 14.74    \\
               & \checkmark  & \checkmark &             &              & -             & -      & 89.45      & 17.05    \\ \hline
    \checkmark & \checkmark  & \checkmark &             &              & 56.71         & 20.51  & 52.04      & 10.06    \\ 
    \checkmark & \checkmark  & \checkmark & \checkmark  &              & 56.18         & 18.72  & 50.35      & 9.31     \\ 
    \checkmark & \checkmark  & \checkmark & \checkmark  & \checkmark   & 56.45         & 19.22  & \textbf{50.22}      & \textbf{8.28}     \\ \hline \hline
    \end{tabular}}
    \caption{Quantitative comparisons on components of DGDM.}
    \label{tab:tab_component}
\end{table}
\begin{figure}[!t]
    \centering
    \includegraphics[width=0.99\linewidth]{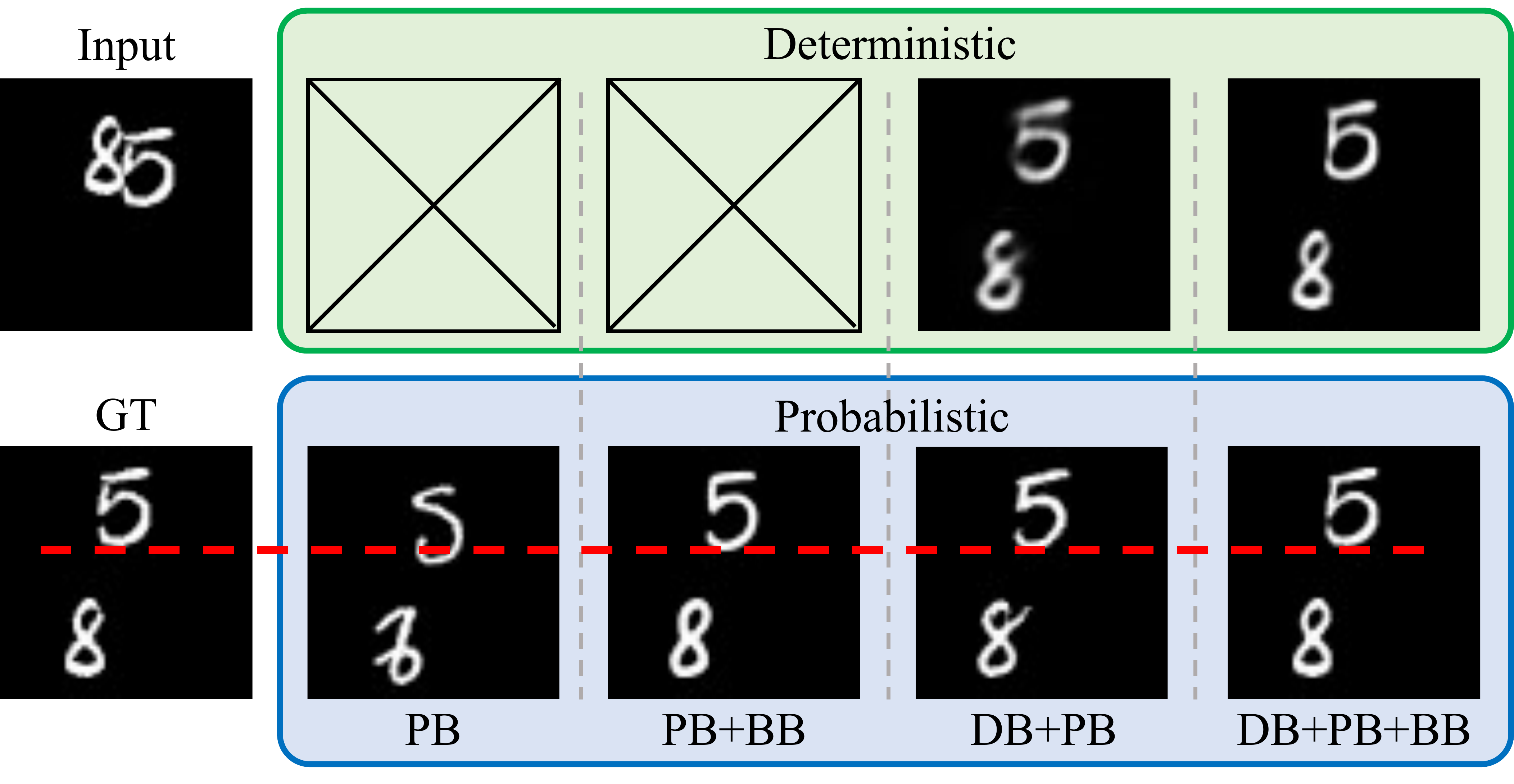}
    \caption{Visualization the results by components. For illustration purposes, we use the same data as in Figure 4.}
    \label{fig:comp}
\end{figure}
\paragraph{Effect of Each Component} 
As shown in \Cref{tab:tab_component} and \Cref{fig:comp}, using only the results of PB in clear images that, however, differ from the GT. Moreover, when PB is combined with the BB, the resulting images are more closely aligned with the actual GT, as opposed to those produced by PB alone.
Lastly, the DB, the replication of the last frame (LF), and the SVS all contribute to improvements in MAE and FVD performance.
These experimental results demonstrate that all components of DGDM are complementary to each other.
Interestingly, it has been observed that when DB and PB are trained together, the performance of DB also improves.
This outcome indicates that the DGDM pipeline is effective not only for PB but also for DB.
Particularly noteworthy is that SVS enhances MAE and FVD performance and also accelerates inference speed for frames with shorter lead times.
While the total diffusion steps may be the same as those in diffusion models without SVS, SVS completes inference earlier for frames with shorter lead times due to fewer diffusion steps being required.
\paragraph{Effect of Truncated Diffusion with Deterministic}
\tableref{tab:ablation_initial} shows the results of the reverse process based on the denoising steps. 
When using a non-Markovian chain without truncated diffusion, there appears to be no significant relationship between the number of diffusion steps and MAE, MSE, and FVD.
On the other hand, setting the reverse step to 200 and using deterministic results to truncate the reverse process to reduce the diffusion step significantly improves accuracy.
In addition, a trend is apparent where reducing diffusion steps leads to a lower standard deviation (STD) of MAE, MSE and FVD.
This suggests that our method can effectively modulate uncertainty in weather forecasting.
In addition, truncated diffusion improves computational efficiency without degrading performance.
%
\begin{table}[t!]
\resizebox{\columnwidth}{!}{%
    \centering
    \begin{tabular}{cc|cc|cc|cc}
    \hline \hline
    \multicolumn{2}{c|}{\multirow{3}{*}{Diffusion steps}} & \multicolumn{5}{c}{\textbf{Evaluate metric}} \\ \cline{3-8} 
                               &     & \multicolumn{2}{c|}{MSE} & \multicolumn{2}{c|}{MAE} & \multicolumn{2}{c}{FVD}  \\ 
                               &     & Mean      & STD      & Mean      & STD      & Mean  & STD          \\ \hline
    \multirow{5}{*}{w/o truncated}  & 200 & 23.621     & 0.105    & 51.223     & 0.126    & 8.282 & 0.087 \\
                               & 175 & 23.326     & 0.066    & 50.561     & 0.079    & 8.600 & 0.121 \\
                               & 150 & 23.146     & 0.066    & 50.363     & 0.083    & 8.221 & 0.149 \\
                               & 125 & 23.453     & 0.145    & 51.340     & 0.190    & 8.487 & 0.122 \\
                               & 100 & 23.222     & 0.058    & 52.046     & 0.081    & 8.132 & 0.154 \\ \hline
    \multirow{4}{*}{w/ truncated} & 175 & 21.947     & 0.046    & 49.543     & 0.080    & 8.326 & 0.079 \\
                               & 150 & 21.613     & 0.031    & \textbf{49.498}     & 0.052    & 8.126 & 0.080 \\
                               & 125 & 21.370     & 0.028    & 49.792     & 0.034    & 7.913 & 0.072 \\
                               & 100 & \textbf{20.963}     & 0.021    & 50.210     & 0.033    & \textbf{7.461} & 0.068 \\
    \hline \hline
    \end{tabular}}
    \caption{Quantitative results with different diffusion steps.}
    \label{tab:ablation_initial}
\end{table}
\section{Conclusion}
\label{sec:conclusion}
In this paper, we introduce the Deterministic Guidance Diffusion Model for Probabilistic Weather Forecasting (DGDM).
DGDM bridges the gap between the accuracy of deterministic models and the diversity of probabilistic models, addressing the limitations of existing data-driven models that face a trade-off between probabilistic forecasting and accuracy.
To validate DGDM in high-resolution weather forecasting, we introduce the Pacific Northwest Windstorm (PNW)-Typhoon weather satellite dataset.
We evaluate DGDM using the Moving MNIST, PNW-Typhoon, and WeatherBench datasets, and it achieves state-of-the-art results in most of the evaluated metrics.
We envision DGDM becoming a cornerstone in the domain of weather forecasting.

\paragraph{Limitations and Future work}
Due to its utilization of deterministic guidance, DGDM can generate samples that are both closely aligned with the future weather conditions and diverse.
However, a challenge persists for DGDM, as it is fundamentally a probabilistic model.
This necessitates the careful selection of a sample from various possibilities to closely match future weather conditions.
We will study methods for selecting the best sample from a variety of possibilities, rather than simply average ensembling.

\clearpage
{\small
\bibliographystyle{ieee_fullname}
\bibliography{egbib}
}
\clearpage
\onecolumn
\setcounter{page}{1}
\setcounter{section}{0}
\renewcommand{\thesection}{\Alph{section}}

{
\centering
    \Large
    \textbf{\thetitle}\\
    \vspace{0.5em}Supplementary Material \\
    \vspace{1.0em}
}

\section{Introduction}
In this supplementary material we provide additional information as follows:
\begin{itemize}
    \item Architecture details of DGDM.
    \item Additional qualitative results.
\end{itemize}

\section{Implementation Details}
In this section, we provide implementation details of the Deterministic Guidance Diffusion Model (DGDM).
~\tabref{tab:hyper} shows the hyperparameter settings used in all of our experiments with DGDM.
Please note that while the number of training epochs differ depending on the training dataset (Moving MNIST: 2000, PNW-Typhoon: 200, WeatherBench: 50), the hyperparameters remain constant across all datasets.
\begin{table}[h!]
\centering
\begin{tabular}{l|c}
\hline \hline
Hyperparameter     & Value             \\ \hline
Optimizer           & Adam              \\
Learning rate of DB    & $1e^{-3}$    \\ 
Learning rate of PB    & $1e^{-4}$     \\ 
$\beta_{1}$          & 0.9               \\
$\beta_{2}$          & 0.999             \\
Weight decay        & 0.0               \\ \hline
Learning rate scheduler           & ReduceLROnPlateau \\
Min learning rate   & $5e^{-6}$         \\
Factor              & 0.5               \\
Patience            & 3,000             \\
Cool down           & 3,000             \\
Threshold           & $1e^{-4}$         \\ \hline
EMA decay           & 0.995             \\
EMA Start step      & 30,000            \\
EMA update interval & 8                 \\
\hline \hline
\end{tabular}
\caption{DGDM training settings. Note that all hyperparameters are applied equally to all datasets.}
\label{tab:hyper}
\end{table}
%

In addition, we present the detailed architecture of the deterministic and probability branches of the DGDM.
As shown in~\tabref{tab:deter}, the Deterministic branch adopts an encoder-translator-decoder structure, a leading approach in recent deterministic video frame prediction~\cite{seo2023implicit, tan2023temporal}.
\tabref{tab:unet3d} presents the architecture details of the probability branch. The probability branch of DGDM employs a 3D-UNet architecture, which is identical to that used in video diffusion models.
%
%
%
\begin{table*}[t!]
    \centering
    \begin{tabular}{l|l|c|c}
    \hline \hline
    Block                                        & Layer                 & Resolution                              & Channels                         \\ \hline \hline
    \textbf{Encoder}                             &                       &                                         &                                  \\ \hline 
    \multirow{3}{*}{2D CNN}                      & Conv3$\times$3        & $H\times W$                             & $C \rightarrow$ 64               \\
                                                 & LayerNorm             &                                         & 64                               \\
                                                 & SiLU                  &                                         & 64                               \\ \hline
    \multirow{3}{*}{2D CNN}                      & Conv3$\times$3        & $H\times W$ $\rightarrow$ $H/2 \times W/2$ & 64                               \\
                                                 & LayerNorm             &                                         & 64                               \\
                                                 & SiLU                  &                                         & 64                               \\ \hline
    \multirow{3}{*}{2D CNN}                      & Conv3$\times$3        & $H/2 \times W/2$                        & 64                               \\
                                                 & LayerNorm             &                                         & 64                               \\
                                                 & SiLU                  &                                         & 64                               \\ \hline
    \multirow{3}{*}{2D CNN}                      & Conv3$\times$3        & $H/2 \times W/2$ $\rightarrow$ $H/4 \times W/4$ & 64                               \\
                                                 & LayerNorm             &                                         & 64                               \\
                                                 & SiLU                  &                                         & 64                               \\ \hline \hline
    \textbf{Translator}                          &                       &                                         & 64 $\rightarrow$ 64$\times T$    \\ \hline
    \multirow{5}{*}{ConvNext$\times$8}           & Conv7$\times$7        & $H/4 \times W/4$                        & 64$\times T$                              \\
                                                 & LayerNorm             &                                         & 64$\times T$                     \\
                                                 & Linear                &                                         & 64$\times T$ $\rightarrow$ 256$\times T$           \\
                                                 & GELU                  &                                         & 256$\times T$                     \\
                                                 & Linear                &                                         & 256$\times T$ $\rightarrow$ 64$\times T$           \\ \hline \hline
    \textbf{Decoder}                             &                       &                                         & 64$\times T$ $\rightarrow$ 64    \\ \hline
    \multirow{3}{*}{2D CNN}                      & Conv3$\times$3        & $H/4 \times W/4$                        & 64 $\rightarrow$ 256             \\
                                                 & LayerNorm             &                                         & 256                              \\
                                                 & SiLU                  &                                         & 256                              \\ \hline
    Upsample                                     & Pixelshuffle          & $H/4 \times W/4$ $\rightarrow$ $H/2 \times W/2$ & 256 $\rightarrow$ 64             \\ \hline
    \multirow{3}{*}{2D CNN}                      & Conv3$\times$3        & $H/2 \times W/2$                        & 64                               \\
                                                 & LayerNorm             &                                         & 64                               \\
                                                 & SiLU                  &                                         & 64                               \\ \hline
    \multirow{3}{*}{2D CNN}                      & Conv3$\times$3        & $H/2 \times W/2$                        & 64 $\rightarrow$ 256             \\
                                                 & LayerNorm             &                                         & 256                              \\
                                                 & SiLU                  &                                         & 256                              \\ \hline
    Upsample                                     & Pixelshuffle          & $H/2 \times W/2$ $\rightarrow$ $H\times W$ & 256 $\rightarrow$ 64             \\ \hline
    \multirow{3}{*}{2D CNN}                      & Conv3$\times$3        & $H\times W$                            & 64 $\rightarrow$ 64              \\
                                                 & LayerNorm             &                                         & 64                               \\
                                                 & SiLU                  &                                         & 64                               \\ \hline
    \multirow{2}{*}{Readout}                     & Reshape               &                                         & 64 $\rightarrow$ 64$\times T$    \\
                                                 & Conv3$\times$3        & $H\times W$                            & 64$\times T$ $\rightarrow C \times T$             \\ 
    \hline \hline
    \end{tabular}
    \caption{Detailed architecture of the deterministic branch. \textbf{Conv3x3} denotes a 2D convolution layer with a 3$\times$3 kernel size, while \textbf{LayerNorm} refers to layer normalization. \textbf{SiLU} is the Sigmoid Linear Unit activation layer~\cite{hendrycks2016gaussian}, and \textbf{Pixelshuffle} is a upsampling layer~\cite{shi2016real} that rearrange channel dimension into the spatial dimension. Throughout the encoder and decoder, the time dimension $T$ is treated as a batch, while in the Translator and the final convolutional layer, it is reshaped into the channel dimension.}
    \label{tab:deter}
\end{table*}

\begin{table*}[t!]
    \centering
    \renewcommand{\arraystretch}{0.9}
    \resizebox{0.55\textwidth}{!}{%
    \begin{tabular}{l|l|c|c}
    \hline \hline
    Block                           & Layer                    & Resolution                               & Channels               \\ \hline
    \multirow{2}{*}{Initial Block}   & Conv7$\times$7           & $H\times W$                              & $C \rightarrow$ 64     \\
                                    & TeAttention              &                                          & 64                     \\ \hline   
    \multirow{2}{*}{Time   MLP}     & Linear                   &                                          & 64 $\rightarrow$ 256                    \\
                                    & GELU                     &                                          & 256                    \\
                                    & Linear                   &                                          & 256                    \\ \hline
    \multirow{2}{*}{Cond Block}     & Conv3$\times$3           & $H/4 \times W/4$                         & 64                     \\
                                    & GroupNorm8               &                                          & 64                     \\
                                    & SiLU                     &                                          & 64                     \\
                                    & Conv3$\times$3           & $H/4 \times W/4$                         & 64                     \\
                                    & GroupNorm8               &                                          & 64                     \\
                                    & SiLU                     &                                          & 64                     \\ \hline \hline
    \multirow{2}{*}{Down Block}     & ResBlock                 & $H\times W$                             & 64                     \\
                                    & ResBlock                 & $H\times W$                             & 64                     \\
                                    & SpAttention              &                                          & 64                     \\
                                    & TeAttention              &                                          & 64                     \\
                                    & Conv4$\times$4           & $H\times W$ $\rightarrow$ $H/2 \times W/2$  & 64                     \\ \hline
    \multirow{2}{*}{Down Block}     & ResBlock                 & $H/2 \times W/2$                             & 64 $\rightarrow$ 128   \\
                                    & ResBlock                 & $H/2 \times W/2$                             & 128                    \\
                                    & SpAttention              &                                          & 128                    \\
                                    & TeAttention              &                                          & 128                    \\
                                    & Conv4$\times$4           & $H/2 \times W/2$ $\rightarrow$ $H/4 \times W/4$  & 128                    \\ \hline
    \multirow{2}{*}{Down Block}     & ResBlock                 & $H/4 \times W/4$                             & 128 $\rightarrow$ 256  \\
                                    & ResBlock                 & $H/4 \times W/4$                             & 256                    \\
                                    & SpAttention              &                                          & 256                    \\
                                    & TeAttention              &                                          & 256                    \\
                                    & Conv4$\times$4           & $H/4 \times W/4$ $\rightarrow$ $H/8 \times W/8$    & 256                    \\ \hline
    \multirow{2}{*}{Down Block}     & ResBlock                 & $H/8 \times W/8$                               & 256 $\rightarrow$ 512  \\
                                    & ResBlock                 & $H/8 \times W/8$                               & 512                    \\
                                    & SpAttention              &                                          & 512                    \\
                                    & TeAttention              &                                          & 512                    \\ \hline \hline
    \multirow{2}{*}{Mid Block}      & ResBlock                 & $H/8 \times W/8$                               & 512                    \\
                                    & SpAttention              &                                          & 512                    \\
                                    & TeAttention              &                                          & 512                    \\
                                    & ResBlock                 & $H/8 \times W/8$                               & 512                    \\ \hline \hline
    \multirow{2}{*}{Up Block}       & ResBlock                 & $H/8 \times W/8$                               & 1024 $\rightarrow$ 256 \\
                                    & ResBlock                 & $H/8 \times W/8$                               & 256                    \\
                                    & SpAttention              &                                          & 256                    \\
                                    & TeAttention              &                                          & 256                    \\
                                    & Deconv4$\times$4         & $H/8 \times W/8$ $\rightarrow$ $H/4 \times W/4$    & 256                    \\ \hline
    \multirow{2}{*}{Up Block}       & ResBlock                 & $H/4 \times W/4$                             & 512 $\rightarrow$ 128  \\
                                    & ResBlock                 & $H/4 \times W/4$                             & 128                    \\
                                    & SpAttention              &                                          & 128                    \\
                                    & TeAttention              &                                          & 128                    \\
                                    & Deconv4$\times$4         & $H/4 \times W/4$ $\rightarrow$ $H/2 \times W/2$  & 128                    \\ \hline
    \multirow{2}{*}{Up Block}       & ResBlock                 & $H/2 \times W/2$                             & 256 $\rightarrow$ 64   \\
                                    & ResBlock                 & $H/2 \times W/2$                             & 64                     \\
                                    & PoAttention              &                                          & 64                     \\
                                    & TeAttention              &                                          & 64                     \\
                                    & Deconv4$\times$4         & $H/2 \times W/2$ $\rightarrow$ $H\times W$  & 64                     \\ \hline
    \multirow{2}{*}{Up Block}       & ResBlock                 & $H\times W$                             & 128 $\rightarrow$ 64   \\
                                    & ResBlock                 & $H\times W$                             & 64                     \\
                                    & SpAttention              &                                          & 64                     \\
                                    & TeAttention              &                                          & 64                     \\ \hline
    \multirow{2}{*}{Out Block}      & ResBlock                 & $H\times W$                             & 192 $\rightarrow$ 64   \\ 
                                    & Conv1$\times$1           & $H\times W$                             & 64 $\rightarrow C$     \\
    \hline \hline
    \end{tabular}%
    }
    \caption{Detailed architecture of the probabilistic branch. \textbf{Conv7$\times$7}, \textbf{Conv3$\times$3}, and \textbf{Conv1$\times$1} are 3D convolutional layers with kernel sizes of 1$\times$7$\times$7, 1$\times$3$\times$3, and 1$\times$1$\times$1, respectively. \textbf{GroupNorm8} is the Group Normalization layer with 8 groups. \textbf{Time MLP} is the block used for embeded the denoising step $t$. \textbf{SiLU} and \textbf{GeLU} are the Sigmoid Linear Unit and Gaussian Error Linear Unit activation layers~\cite{hendrycks2016gaussian}, respectively. \textbf{ResNetBlock} consists of a sequence with a Linear layer addressing the denoising step $t$ followed by a SiLU activation, two blocks of Conv3$\times$3, GroupNorm8, and SiLU activation. \textbf{SpAttention} is a spatial attention layer that considers only the spatial dimension with deterministic features. \textbf{TeAttention} is a self-attention layer that focuses on the temporal dimension of sequential data. Downsampling is performed by \textbf{Conv4$\times$4}, a 3D convolutional layer with a stride of 2 and a 1$\times$4$\times$4 kernel size and upsampling is performed through \textbf{Deconv4$\times$4}, a 3D transposed convolutional layer with the same stride and kernel size. \textbf{Up block} receives concatenated inputs from the features of previous block and the features of down block, which have the same resolution. The \textbf{Out block} also use skip connection and further concatenates deterministic features for input.}
    \label{tab:unet3d}
\end{table*}

\section{Additional Qualitative Results}
\begin{figure*}[t!]
 \centering
 \includegraphics[width=0.8\linewidth]{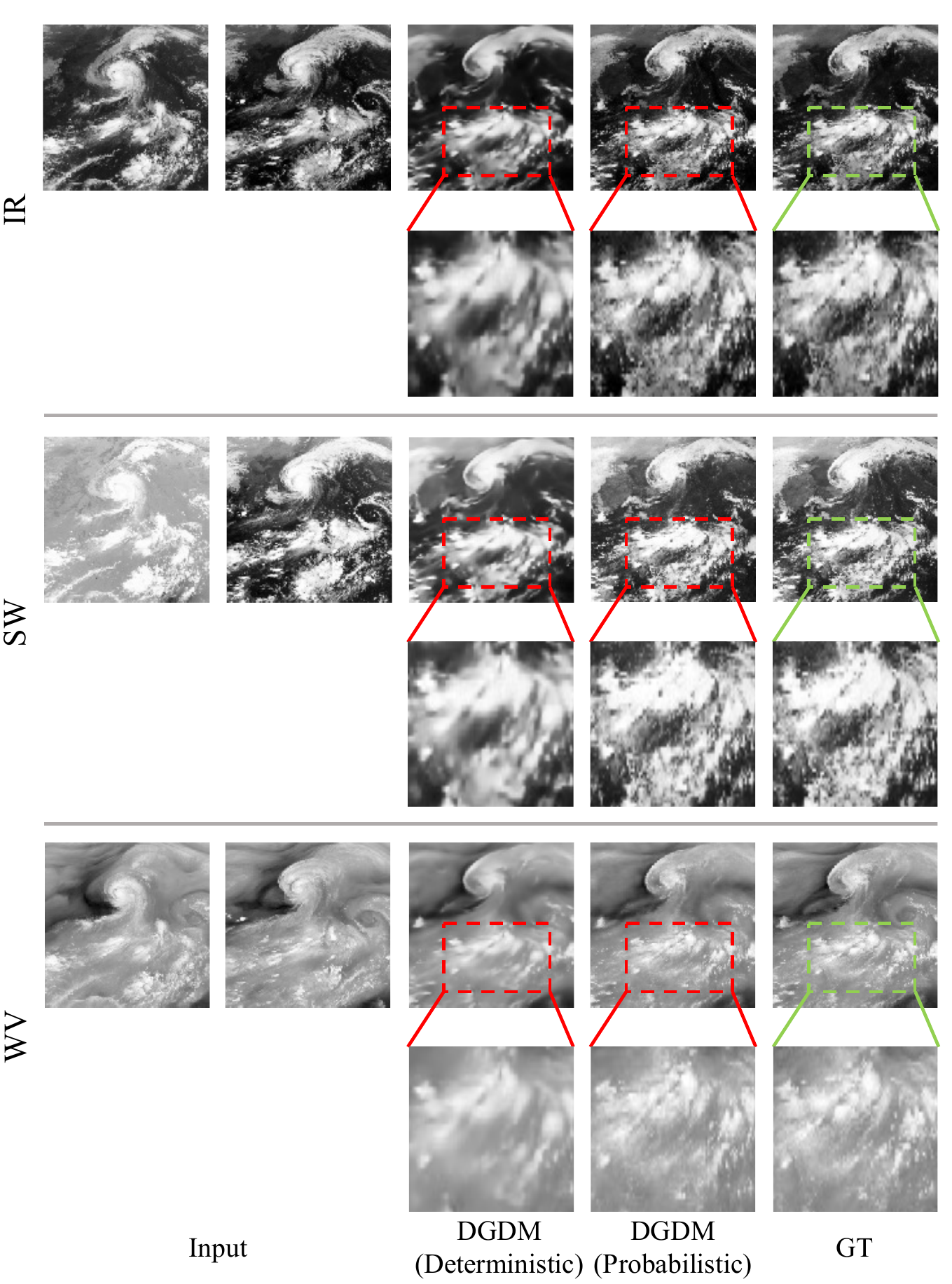}
 \caption{Qualitative experimental results from the PNW-Typhoon dataset with DGDM. The two samples on the left are the 0th and 9th samples, and a total of 10 frames from 0 to 9 are inputted. Also, the results all display the 19th frame, which is 10 hours later.}
 \label{fig:pwn_sup}
\end{figure*}

In this section, we provide additional qualitative results that could not be included in the main text.
~\figureref{fig:pwn_sup} shows the qualitative analysis of DGDM-Best (20 samples) from the PNW-Typhoon dataset. As seen in the figure, DGDM-Deterministic shows blurry results, but DGDM-Probability yields relatively clear results.
These qualitative experimental results demonstrate that DGDM can be useful not only in predicting large-scale climate phenomena but also smaller-scale ones.

\begin{figure*}[t!]
 \centering
 \includegraphics[width=1.0\linewidth]{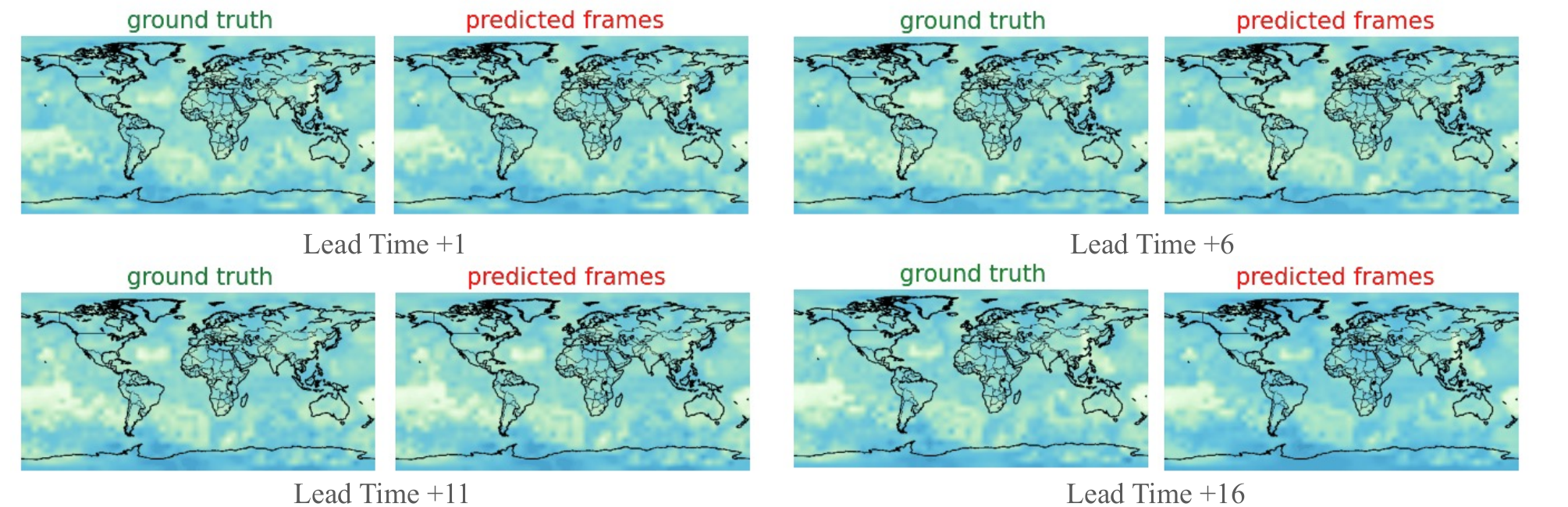}
 \caption{The qualitative results of DGDM from the WeatherBench dataset.}
 \label{fig:era22}
\end{figure*}

~\figureref{fig:era22} shows the qualitative experimental results of DGDM in the WeatherBench dataset.
As can be seen in the figure, DGDM demonstrates its effectiveness in global climate modeling, such as with WeatherBench.


\end{document}